# Bounds Arc Consistency for Weighted CSPs

**Matthias Zytnicki**                    Matthias.Zytnicki@versailles.inra.fr
*INRA, Unité de Recherche en Génomique et Informatique*
*UR 1164, Versailles, France*

**Christine Gaspin**                    Christine.Gaspin@toulouse.inra.fr
**Simon de Givry**                    Simon.DeGivry@toulouse.inra.fr
**Thomas Schiex**                    Thomas.Schiex@toulouse.inra.fr
*INRA, Unité de Biométrie et Intelligence Artificielle*
*UR 875, Toulouse, France*

## Abstract

The Weighted Constraint Satisfaction Problem (WCSP) framework allows representing and solving problems involving both hard constraints and cost functions. It has been applied to various problems, including resource allocation, bioinformatics, scheduling, etc. To solve such problems, solvers usually rely on branch-and-bound algorithms equipped with local consistency filtering, mostly soft arc consistency. However, these techniques are not well suited to solve problems with very large domains. Motivated by the resolution of an RNA gene localization problem inside large genomic sequences, and in the spirit of bounds consistency for large domains in crisp CSPs, we introduce soft *bounds arc consistency*, a new weighted local consistency specifically designed for WCSP with very large domains. Compared to soft arc consistency, BAC provides significantly improved time and space asymptotic complexity. In this paper, we show how the semantics of cost functions can be exploited to further improve the time complexity of BAC. We also compare both in theory and in practice the efficiency of BAC on a WCSP with bounds consistency enforced on a crisp CSP using cost variables. On two different real problems modeled as WCSP, including our RNA gene localization problem, we observe that maintaining bounds arc consistency outperforms arc consistency and also improves over bounds consistency enforced on a constraint model with cost variables.

## 1. Introduction

The Weighted Constraint Satisfaction Problem (WCSP) is an extension of the crisp Constraint Satisfaction Problem (CSP) that allows the direct representation of hard constraints and cost functions. The WCSP defines a simple optimization (minimization) framework with a wide range of applications in *resource allocation*, *scheduling*, *bioinformatics* (Sànchez, de Givry, & Schiex, 2008; Zytnicki, Gaspin, & Schiex, 2008), *electronic markets* (Sandholm, 1999), etc. It also captures fundamental AI and statistical problems such as Maximum Probability Explanation in Bayesian nets and Markov Random Fields (Chellappa & Jain, 1993).

As in crisp CSP, the two main approaches to solve WCSP are inference and search. This last approach is usually embodied in a branch-and-bound algorithm. This algorithm estimates at each node of the search tree a lower bound of the cost of the solutions of the sub-tree.

One of the most successful approaches to build lower bounds has been obtained by extending the notion of local consistency to WCSP (Meseguer, Rossi, & Schiex, 2006). This includes soft AC (Schiex, 2000), AC* (Larrosa, 2002), FDAC* (Larrosa & Schiex, 2004), EDAC* (Heras, Larrosa, de Givry, & Zytnicki, 2005), OSAC (Cooper, de Givry, & Schiex, 2007) and VAC (Cooper, de Givry, Sànchez, Schiex, & Zytnicki, 2008) among others. Unfortunately, the worst case time complexity bounds of the associated enforcing algorithms are at least cubic in the domain size and use an amount of space which is at least linear in the domain size. This makes these consistencies useless for problems with very large domains.

The motivation for designing a local consistency which can be enforced efficiently on problems with large domains follows from our interest in the RNA gene localization problem. Initially modeled as a crisp CSP, this problem has been tackled using bounds consistency (Choi, Harvey, Lee, & Stuckey, 2006; Lhomme, 1993) and dedicated propagators using efficient pattern matching algorithms (Thébault, de Givry, Schiex, & Gaspin, 2006). The domain sizes are related to the size of the genomic sequences considered and can reach hundreds of millions of values. In order to enhance this tool with scoring capabilities and improved quality of localization, a shift from crisp to weighted CSP is a natural step which requires the extension of bounds consistency to WCSP. Beyond this direct motivation, this extension is also useful in other domains where large domains occur naturally such as temporal reasoning or scheduling.

The local consistencies we define combine the principles of bounds consistency with the principles of soft local consistencies. These definitions are general and are not restricted to binary cost functions. The corresponding enforcing algorithms improve over the time and space complexity of AC* by a factor of $d$ and also have the nice but rare property, for WCSP local consistencies, of being confluent.

As it has been done for AC-5 by Van Hentenryck, Deville, and Teng (1992) for functional or monotonic constraints, we show that different forms of cost functions (largely captured by the notion of semi-convex cost functions) can be processed more efficiently. We also show that the most powerful of these bounds arc consistencies is strictly stronger than the application of bounds consistency to the reified representation of the WCSP as proposed by Petit, Régin, and Bessière (2000).

To conclude, we experimentally compare the efficiency of algorithms that maintain these different local consistencies inside branch-and-bound on agile satellite scheduling problems (Verfaillie & Lemaître, 2001) and RNA gene localization problems (Zytnicki et al., 2008) and observe clear speedups compared to different existing local consistencies.

## 2. Definitions and Notations

This section will introduce the main notions that will be used throughout the paper. We will define the (Weighted) Constraint Satisfaction Problems, as well as a local consistency property frequently used for solving the Weighted Constraint Satisfaction Problem: arc consistency (AC*).

## 2.1 Constraint Networks

Classic and weighted constraint networks share finite domain variables as one of their components. In this paper, the domain of a variable $x_i$ is denoted by $D(x_i)$. To denote a value in $D(x_i)$, we use an index $i$ as in $v_i$, $v'_i$,... For each variable $x_i$, we assume that the domain of $x_i$ is totally ordered by $\prec_i$ and we denote by $\inf(x_i)$ and $\sup(x_i)$ the minimum (resp. maximum) values of the domain $D(x_i)$. An *assignment* $t_S$ of a set of variables $S = \{x_{i_1}, \ldots, x_{i_r}\}$ is a function that maps variables to elements of their domains: $t_S = (x_{i_1} \leftarrow v_{i_1}, \ldots, x_{i_r} \leftarrow v_{i_r})$ with $\forall i \in \{i_1, \ldots, i_r\}, t_S(x_i) = v_i \in D(x_i)$. For a given assignment $t_S$ such that $x_i \in S$, we simply say that a value $v_i \in D(x_i)$ belongs to $t_S$ to mean that $t_S(x_i) = v_i$. We denote by $\ell_S$, the set of all possible assignments on $S$.

**Definition 2.1** *A constraint network (CN) is a tuple $\mathcal{P} = \langle \mathcal{X}, \mathcal{D}, \mathcal{C} \rangle$, where $\mathcal{X} = \{x_1, \ldots, x_n\}$ is a set of variables and $\mathcal{D} = \{D(x_1), \ldots, D(x_n)\}$ is the set of the finite domains of each variable. $\mathcal{C}$ is a set of constraints. A constraint $c_S \in \mathcal{C}$ defines the set of all authorized combinations of values for the variables in $S$ as a subset of $\ell_S$. $S$ is called the scope of $c_S$.*

$|S|$ is called the arity of $c_S$. For simplicity, unary (arity 1) and binary (arity 2) constraints may be denoted by $c_i$ and $c_{ij}$ instead of $c_{\{x_i\}}$ and $c_{\{x_i, x_j\}}$ respectively. We denote by $d$ the maximum domain size, $n$, the number of variables in the network and $e$, the number of constraints. The central problem on constraint networks is to find a solution, defined as an assignment $t_{\mathcal{X}}$ of all variables such that for any constraint $c_S \in \mathcal{C}$, the restriction of $t_{\mathcal{X}}$ to $S$ is authorized by $c_S$ (all constraints are satisfied). This is the Constraint Satisfaction Problem (CSP).

**Definition 2.2** *Two CNs with the same variables are equivalent if they have the same set of solutions.*

A CN will be said to be "empty" if one of its variables has an empty domain. This may happen following local consistency enforcement. For CN with large domains, the use of bounds consistency is the most usual approach. Historically, different variants of bounds consistency have been introduced, generating some confusion. Using the terminology introduced by Choi et al. (2006), the bounds consistency considered in this paper is the $bounds(\mathcal{D})$ consistency. Because we only consider large domains defining intervals, this is actually equivalent to $bounds(\mathcal{Z})$ consistency. For simplicity, in the rest of the paper we denote this as "bounds consistency".

**Definition 2.3 (Bounds consistency)** *A variable $x_i$ is bounds consistent iff every constraint $c_S \in \mathcal{C}$ such that $x_i \in S$ contains a pair of assignments $(t, t') \in \ell_S \times \ell_S$ such that $\inf(x_i) \in t$ and $\sup(x_i) \in t'$. In this case, $t$ and $t'$ are called the supports of the two bounds of $x_i$'s domain.*

*A CN is bounds consistent iff all its variables are bounds consistent.*

To enforce bounds consistency on a given CN, any domain bound that does not satisfy the above properties is deleted until a fixed point is reached.

## 2.2 Weighted Constraint Networks

Weighted constraint networks are obtained by using cost functions (also referred as "soft constraints") instead of constraints.

**Definition 2.4** *A weighted constraint network (WCN) is a tuple $\mathcal{P} = \langle \mathcal{X}, \mathcal{D}, \mathcal{W}, k \rangle$, where $\mathcal{X} = \{x_1, \ldots, x_n\}$ is a set of variables and $\mathcal{D} = \{D(x_1), \ldots, D(x_n)\}$ is the set of the finite domains of each variable. $\mathcal{W}$ is a set of cost functions. A cost function $w_S \in \mathcal{W}$ associates an integer cost $w_S(t_S) \in [0, k]$ to every assignment $t_S$ of the variables in $S$. The positive number $k$ defines a maximum (intolerable) cost.*

The cost $k$, which may be finite or infinite, is the cost associated with forbidden assignments. This cost is used to represent hard constraints. Unary and binary cost functions may be denoted by $w_i$ and $w_{ij}$ instead of $w_{\{x_i\}}$ and $w_{\{x_i, x_j\}}$ respectively. As usually for WCNs, we assume the existence of a zero-arity cost function, $w_\varnothing \in [0, k]$, a constant cost whose initial value is usually equal to 0. The cost of an assignment $t_\mathcal{X}$ of all variables is obtained by combining the costs of all the cost functions $w_S \in W$ applied to the restriction of $t_\mathcal{X}$ to $S$. The combination is done using the function $\oplus$ defined as $a \oplus b = \min(k, a + b)$.

**Definition 2.5** *A solution of a WCN is an assignment $t_\mathcal{X}$ of all variables whose cost is less than $k$. It is optimal if no other assignment of $\mathcal{X}$ has a strictly lower cost.*

The central problem in WCN is to find an optimal solution.

**Definition 2.6** *Two WCNs with the same variables are equivalent if they give the same cost to any assignments of all their variables.*

Initially introduced by Schiex (2000), the extension of arc consistency to WCSP has been refined by Larrosa (2002) leading to the definition of AC*. It can be decomposed into two sub-properties: node and arc consistency itself.

**Definition 2.7 (Larrosa, 2002)** *A variable $x_i$ is node consistent iff:*

- *$\forall v_i \in D(x_i), w_\varnothing \oplus w_i(v_i) < k$.*

- *$\exists v_i \in D(x_i)$ such that $w_i(v_i) = 0$. The value $v_i$ is called the unary support of $x_i$.*

*A WCN is node consistent iff every variable is node consistent.*

To enforce NC on a WCN, values that violate the first property are simply deleted. Value deletion alone is not capable of enforcing the second property. As shown by Cooper and Schiex (2004), the fundamental mechanism required here is the ability to move costs between different scopes. A cost $b$ can be subtracted from a greater cost $a$ by the function $\ominus$ defined by $a \ominus b = (a - b)$ if $a \neq k$ and $k$ otherwise. Using $\ominus$, a unary support for a variable $x_i$ can be created by subtracting the smallest unary cost $\min_{v_i \in D(x_i)} w_i(v_i)$ from all $w_i(v_i)$ and adding it (using $\oplus$) to $w_\varnothing$. This operation that shifts costs from variables to $w_\varnothing$, creating a unary support, is called a *projection* from $w_i$ to $w_\varnothing$. Because $\ominus$ and $\oplus$ cancel out, defining a *fair* valuation structure (Cooper & Schiex, 2004), the obtained WCN is equivalent to the original one. This equivalence preserving transformation (Cooper and Schiex) is more precisely described as the ProjectUnary() function in Algorithm 1.

We are now able to define arc and AC* consistency on WCN.

---

**Algorithm 1**: Projections at unary and binary levels

**1 Procedure** ProjectUnary($x_i$)                    [ Find the unary support of $x_i$ ]
**2**  $\quad min \leftarrow \min_{v_i \in D(x_i)}\{w_i(v_i)\}$ ;
**3**  $\quad$ **if** $(min = 0)$ **then return**;
**4**  $\quad$ **foreach** $v_i \in D(x_i)$ **do** $w_i(v_i) \leftarrow w_i(v_i) \ominus min$ ;
**5**  $\quad w_\varnothing \leftarrow w_\varnothing \oplus min$ ;

**6 Procedure** Project($x_i, v_i, x_j$)                    [ Find the support of $v_i$ w.r.t. $w_{ij}$ ]
**7**  $\quad min \leftarrow \min_{v_j \in D(x_j)}\{w_{ij}(v_i, v_j)\}$ ;
**8**  $\quad$ **if** $(min = 0)$ **then return**;
**9**  $\quad$ **foreach** $v_j \in D(x_j)$ **do** $w_{ij}(v_i, v_j) \leftarrow w_{ij}(v_i, v_j) \ominus min$ ;
**10** $\quad w_i(v_i) \leftarrow w_i(v_i) \oplus min$ ;

---

**Definition 2.8** *A variable $x_i$ is arc consistent iff for every cost function $w_S \in \mathcal{W}$ such that $x_i \in S$, and for every value $v_i \in D(x_i)$, there exists an assignment $t \in \ell_S$ such that $v_i \in t$ and $w_S(t) = 0$. The assignment $t$ is called the support of $v_i$ on $w_S$. A WCN is AC\* iff every variable is arc and node consistent.*

To enforce arc consistency, a support for a given value $v_i$ of $x_i$ on a cost function $w_S$ can be created by subtracting (using $\ominus$) the cost $\min_{t \in \ell_S, v_i \in t} w_S(t)$ from the costs of all assignments containing $v_i$ in $\ell_S$ and adding it to $w_i(v_i)$. These cost movements, applied for all values $v_i$ of $D(x_i)$, define the projection from $w_S$ to $w_i$. Again, this transformation preserves equivalence between problems. It is more precisely described (for simplicity, in the case of binary cost functions) as the Project() function in Algorithm 1.

**Example 2.9** Consider the WCN in Figure 1(a). It contains two variables ($x_1$ and $x_2$), each with two possible values ($a$ and $b$, represented by vertices). A unary cost function is associated with each variable, the cost of a value being represented inside the corresponding vertex. A binary cost function between the two variables is represented by weighted edges connecting pairs of values. The absence of edge between two values represents a zero cost. Assume $k$ is equal to 4 and $w_\varnothing$ is equal to 0.

Since the cost $w_1(x_1 \leftarrow a)$ is equal to $k$, the value $a$ can be deleted from the domain of $x_1$ (by NC, first property). The resulting WCN is represented in Figure 1(b). Then, since $x_2$ has no unary support (second line of the definition of NC), we can project a cost of 1 to $w_\varnothing$ (cf. Figure 1(c)). The instance is now NC. To enforce AC\*, we project 1 from the binary cost function $w_{12}$ to the value $a$ of $x_1$ since this value has no support on $w_{12}$ (cf. Figure 1(d)). Finally, we project 1 from $w_1$ to $w_\varnothing$, as seen on Figure 1(e). Ultimately, we note that the value $b$ of $x_2$ has no support. To enforce AC\*, we project a binary cost of 1 to this value and remove it since it has a unary cost of 2 which, combined with $w_\varnothing$ reaches $k = 4$.

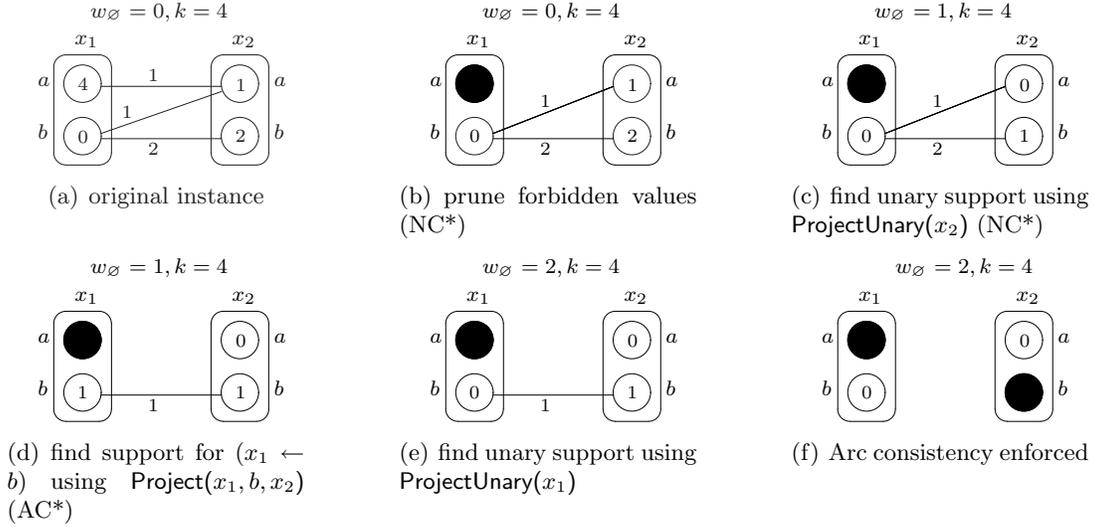

Figure 1: Enforcing Arc Consistency.

## 3. Bounds Arc Consistency (BAC)

In crisp CSP, the bounds consistency enforcing process just deletes bounds that are not supported in one constraint. In weighted CSP, enforcement is more complex. If a similar value deletion process exists based on the first node consistency property violation (whenever $w_\varnothing \oplus w_i(v_i)$ reaches $k$), additional cost movements are performed to enforce node and arc consistency.

As shown for AC*, these projections require the ability to represent an arbitrary unary cost function $w_i$ for every variable $x_i$. This requires space in $O(d)$ in general since projections can lead to arbitrary changes in the original $w_i$ cost function (even if they have an efficient internal representation). To prevent this, we therefore avoid to move cost from cost functions with arity greater than one to unary constraints. Instead of such projections, we only keep a value deletion mechanism applied to the bounds of the current domain that takes into account all the cost functions involving the variable considered. For a given variable $x_i$ involved in a cost function $w_S$, the choice of a given value $v_i$ will at least induce a cost increase of $\min_{t_S \in \ell_S, v_i \in t_S} w_S(t_S)$. If these minimum costs, combined on all the cost functions involving $x_i$, together with $w_\varnothing$, reach the intolerable cost of $k$, then the value can be deleted. As in bounds consistency, this is just done for the two bounds of the domain. This leads to the following definition of BAC (bounds arc consistency) in WCSP:

**Definition 3.1** *In a WCN $P = \langle \mathcal{X}, \mathcal{D}, \mathcal{W}, k \rangle$, a variable $x_i$ is bounds arc consistent iff:*

$$w_\varnothing \oplus \sum_{w_S \in \mathcal{W}, x_i \in S} \left\{ \min_{t_S \in \ell_S, \inf(x_i) \in t_S} w_S(t_S) \right\} < k$$

$$w_\varnothing \oplus \sum_{w_S \in \mathcal{W}, x_i \in S} \left\{ \min_{t_S \in \ell_S, \sup(x_i) \in t_S} w_S(t_S) \right\} < k$$

*A WCN is bounds arc consistent if every variable is bounds arc consistent.*

One can note that this definition is a proper generalization of bounds consistency since when $k = 1$, it is actually equivalent to the definition of $bounds(\mathcal{D})$ consistency for crisp CSP (Choi et al., 2006) (also equivalent to $bounds(\mathcal{Z})$ consistency since domains are defined as intervals).

The algorithm enforcing BAC is described as Algorithm 2. Because enforcing BAC only uses value deletion, it is very similar in structure to bounds consistency enforcement. We maintain a queue $Q$ of variables whose domain has been modified (or is untested). For better efficiency, we use extra data-structures to efficiently maintain the combined cost associated with the domain bound $\inf(x_i)$, denoted $w^{\inf}(x_i)$. For a cost function $w_S$ involving $x_i$, the contribution of $w_S$ to this combined cost is equal to $\min_{t_S \in \ell_S, \inf(x_i) \in t_S} w_S(t_S)$. This contribution is maintained in a data-structure $\Delta^{\inf}(x_i, w_S)$ and updated whenever the minimum cost may change because of value removals. Notice that, in Algorithm 2, the line 14 is a concise way to denote the hidden loops which initialize the $w^{\inf}$, $w^{\sup}$, $\Delta^{\inf}$ and $\Delta^{\sup}$ data-structures to zero.

Domain pruning is achieved by function PruneInf() which also resets the data-structures associated with the variable at line 35 and these data-structures are recomputed when the variable is extracted from the queue. Indeed, inside the loop of line 20, the contributions $\Delta^{\inf}(x_i, w_S)$ to the cost $w^{\inf}(x_i)$ from the cost functions $w_S$ involving $x_j$ are reset. The Function *pop* removes an element from the queue and returns it.

**Proposition 3.2 (Time and space complexity)** *For a WCN with maximum arity $r$ of the constraints, enforcing BAC with Algorithm 2 is time $O(er^2d^r)$ and space $O(n + er)$.*

**Proof:** Regarding time, every variable can be pushed into $Q$ at most $d + 1$ times: once at the beginning, and when one of its values has been removed. As a consequence, the **foreach** loop on line 18 iterates $O(erd)$ times, and the **foreach** loop on line 20 iterates $O(er^2d)$ times. The min computation on line 22 takes time $O(d^{r-1})$ and thus, the overall time spent at this line takes time $O(er^2d^r)$. PruneInf() is called at most $O(er^2d)$ times. The condition on line 32 is true at most $O(nd)$ times and so, line 35 takes time $O(ed)$ (resetting $\Delta^{\inf}(x_i, \cdot)$ on line 35 hides a loop on all cost functions involving $x_i$). The total time complexity is thus $O(er^2d^r)$.

Regarding space, we only used $w^{\inf}$, $w^{\sup}$ and $\Delta$ data-structures. The space complexity is thus $O(n + er)$. $\qquad\square$

Note that exploiting the information of last supports as in AC2001 (Bessière & Régin, 2001) does not reduce the worst-case time complexity because the minimum cost of a cost function must be recomputed from scratch each time a domain has been reduced and the last support has been lost (Larrosa, 2002). However, using last supports helps in practice to reduce mean computation time and this has been done in our implementation.

Compared to AC*, which can be enforced in $O(n^2d^3)$ time and $O(ed)$ space for binary WCN, BAC can be enforced $d$ times faster, and the space complexity becomes independent of $d$ which is a requirement for problems with very large domains.

Another interesting difference with AC* is that BAC is confluent — just as bounds consistency is. Considering AC*, it is known that there may exist several different AC* closures with possibly different associated lower bounds $w_\varnothing$ (Cooper & Schiex, 2004). Note that although OSAC (Cooper et al., 2007) is able to find an optimal $w_\varnothing$ (at much higher

---

**Algorithm 2**: Algorithm enforcing BAC.

---

11 **Procedure** $\mathsf{BAC}(\mathcal{X}, \mathcal{D}, \mathcal{W}, k)$

12     $Q \leftarrow \mathcal{X}$ ;

14     $w^{\inf}(\cdot) \leftarrow 0$ ; $w^{\sup}(\cdot) \leftarrow 0$ ; $\Delta^{\inf}(\cdot, \cdot) \leftarrow 0$ ; $\Delta^{\sup}(\cdot, \cdot) \leftarrow 0$ ;

15     **while** $(Q \neq \varnothing)$ **do**

16        $x_j \leftarrow pop(Q)$ ;

18        **foreach** $w_S \in \mathcal{W}, x_j \in S$ **do**

20           **foreach** $x_i \in S$ **do**

22              $\alpha \leftarrow \min_{t_S \in \ell_S, \inf(x_i) \in t_S} w_S(t_S)$ ;

23              $w^{\inf}(x_i) \leftarrow w^{\inf}(x_i) \ominus \Delta^{\inf}(x_i, w_S) \oplus \alpha$ ;

24              $\Delta^{\inf}(x_i, w_S) \leftarrow \alpha$ ;

25              **if** $\mathsf{PruneInf}(x_i)$ **then** $Q \leftarrow Q \cup \{x_i\}$ ;

26              $\alpha \leftarrow \min_{t_S \in \ell_S, \sup(x_i) \in t_S} w_S(t_S)$ ;

27              $w^{\sup}(x_i) \leftarrow w^{\sup}(x_i) \ominus \Delta^{\sup}(x_i, w_S) \oplus \alpha$ ;

28              $\Delta^{\sup}(x_i, w_S) \leftarrow \alpha$ ;

29              **if** $\mathsf{PruneSup}(x_i)$ **then** $Q \leftarrow Q \cup \{x_i\}$ ;

30 **Function** $\mathsf{PruneInf}(x_i)$ *: boolean*

32     **if** $(w_\varnothing \oplus w^{\inf}(x_i) = k)$ **then**

33        delete $\inf(x_i)$ ;

35        $w^{\inf}(x_i) \leftarrow 0$ ; $\Delta^{\inf}(x_i, \cdot) \leftarrow 0$ ;

36        **return** true;

37     **else return** false;

38 **Function** $\mathsf{PruneSup}(x_i)$ *: boolean*

39     **if** $(w_\varnothing \oplus w^{\sup}(x_i) = k)$ **then**

40        delete $\sup(x_i)$ ;

41        $w^{\sup}(x_i) \leftarrow 0$ ; $\Delta^{\sup}(x_i, \cdot) \leftarrow 0$ ;

42        **return** true;

43     **else return** false;

---

computational cost), it is still not confluent. The following property shows that BAC is confluent.

**Proposition 3.3 (Confluence)** *Enforcing BAC on a given problem always leads to a unique WCN.*

**Proof:** We will prove the proposition as follows. We will first define a set of problems which contains all the problems that can be reached from the original WCN through BAC enforcement. Notice that, at each step of BAC enforcement, in the general case, several operations can be performed and no specific order is imposed. Therefore, a set of problems can be reached at each step. We will show that the set of problems has a lattice structure and ultimately show that the closure of BAC is the lower bound of this lattice, and is therefore unique, which proves the property. This proof technique is usual for proving convergence of the chaotic iteration of a collection of suitable functions and has been used for characterizing CSP local consistency by Apt (1999).

During the enforcement of BAC, the original problem $\mathcal{P} = \langle \mathcal{X}, \mathcal{D}, \mathcal{W}, k \rangle$ is iteratively transformed into a set of different problems which are all equivalent to $\mathcal{P}$, and obtained by deleting values violating BAC. Because these problems are obtained by value removals, they belong to the set $\wp_1(P)$ defined by: $\{\langle \mathcal{X}, \mathcal{D}', \mathcal{W}, k \rangle : \mathcal{D}' \subseteq \mathcal{D}\}$.

We now define a relation, denoted $\sqsubseteq$, on the set $\wp_1(P)$:

$$\forall (\mathcal{P}_1, \mathcal{P}_2) \in \wp_1^2(\mathcal{P}), \mathcal{P}_1 \sqsubseteq \mathcal{P}_2 \Leftrightarrow \forall i \in [1, n], D_1(x_i) \subseteq D_2(x_i)$$

It is easy to see that this relation defines a partial order. Furthermore, each pair of elements has a greatest lower bound glb and a least upper bound lub in $\wp_1(\mathcal{P})$, defined by:

$$\forall (\mathcal{P}_1, \mathcal{P}_2) \in \wp_1^2(\mathcal{P}),$$
$$\mathrm{glb}(\mathcal{P}_1, \mathcal{P}_2) = \langle \mathcal{X}, \{D_1(x_i) \cap D_2(x_i) : i \in [1, n]\}, \mathcal{W}, k \rangle \in \wp_1(\mathcal{P})$$
$$\mathrm{lub}(\mathcal{P}_1, \mathcal{P}_2) = \langle \mathcal{X}, \{D_1(x_i) \cup D_2(x_i) : i \in [1, n]\}, \mathcal{W}, k \rangle \in \wp_1(\mathcal{P})$$

$\langle \wp_1(\mathcal{P}), \sqsubseteq \rangle$ is thus a complete lattice.

BAC filtering works by removing values violating the BAC properties, transforming an original problem into a succession of equivalent problems. Each transformation can be described by the application of dedicated functions from $\wp_1(\mathcal{P})$ to $\wp_1(\mathcal{P})$. More precisely, there are two such functions for each variable, one for the minimum bound $\inf(x_i)$ of the domain of $x_i$ and a symmetrical one for the maximum bound. For $\inf(x_i)$, the associated function keeps the instance unchanged if $\inf(x_i)$ satisfies the condition of Definition 3.1 and it otherwise returns a WCN where $\inf(x_i)$ alone has been deleted. The collection of all those functions defines a set of functions from $\wp_1(P)$ to $\wp_1(P)$ which we denote by $F_{BAC}$.

Obviously, every function $f \in F_{BAC}$ is order preserving:

$$\forall (\mathcal{P}_1, \mathcal{P}_2) \in \wp_1^2(\mathcal{P}), \mathcal{P}_1 \sqsubseteq \mathcal{P}_2 \Rightarrow f(\mathcal{P}_1) \sqsubseteq f(\mathcal{P}_2)$$

By application of the Tarski-Knaster theorem (Tarski, 1955), it is known that every function $f \in F_{BAC}$ (applied until quiescence during BAC enforcement) has at least one fixpoint, and that the set of these fixed points forms a lattice for $\sqsubseteq$. Moreover, the intersection of the lattices of fixed points of the functions $f \in F_{BAC}$, denoted by $\wp_1^\star(\mathcal{P})$, is also a lattice. $\wp_1^\star(\mathcal{P})$ is not empty since the problem $\langle \mathcal{X}, \{\varnothing, \ldots, \varnothing\}, \mathcal{W} \rangle$ is a fixpoint for every filtering function in $F_{BAC}$. $\wp_1^\star(\mathcal{P})$ is exactly the set of fixed points of $F_{BAC}$.

We now show that, if the algorithm reaches a fixpoint, it reaches the greatest element of $\wp_1^\star(\mathcal{P})$. We will prove by induction that any successive application of elements of $F_{BAC}$ on $\mathcal{P}$ yields problems which are greater than any element of $\wp_1^\star(\mathcal{P})$ for the order $\sqsubseteq$. Let us consider any fixpoint $\mathcal{P}^\star$ of $\wp_1^\star(\mathcal{P})$. Initially, the algorithm applies on $\mathcal{P}$, which is the greatest element of $\wp_1(\mathcal{P})$, and thus $\mathcal{P}^\star \sqsubseteq \mathcal{P}$. This is the base case of the induction. Let us now consider any problem $\mathcal{P}_1$ obtained during the execution of the algorithm. We have, by induction, $\mathcal{P}^\star \sqsubseteq \mathcal{P}_1$. Since $\sqsubseteq$ is order preserving, we know that, for any function $f$ of $F_{BAC}$, $f(\mathcal{P}^\star) = \mathcal{P}^\star \sqsubseteq f(\mathcal{P}_1)$. This therefore proves the induction.

To conclude, if the algorithm terminates, then it gives the maximum element of $\wp_1^\star(\mathcal{P})$. Since proposition 3.2 showed that the algorithm actually terminates, we can conclude that it is confluent. $\qquad\square$

If enforcing BAC may reduce domains, it never increases the lower bound $w_\varnothing$. This is an important limitation given that each increase in $w_\varnothing$ may generate further value deletions and possibly, failure detection. Note that even when a cost function becomes totally assigned, the cost of the corresponding assignment is not projected to $w_\varnothing$ by BAC enforcement. This can be simply done by maintaining a form of backward checking as in the most simple WCSP branch-and-bound algorithm (Freuder & Wallace, 1992). To go beyond this simple approach, we consider the combination of BAC with another WCSP local consistency which, similarly to AC*, requires cost movements to be enforced but which avoids the modification of unary cost functions to keep a reasonable space complexity. This is achieved by directly moving costs to $w_\varnothing$.

## 4. Enhancing BAC

In many cases, BAC may be very weak compared to AC* in situations where it seems to be possible to infer a decent $w_\varnothing$ value. Consider for example the following cost function:

$$w_{12} : \left\{ \begin{array}{ccc} D(x_1) \times D(x_2) & \to & E \\ (v_1, v_2) & \mapsto & v_1 + v_2 \end{array} \right. \qquad D(x_1) = D(x_2) = [1, 10]$$

AC* can increase $w_\varnothing$ by 2, by projecting a cost of 2 from $w_{12}$ to the unary constraint $w_1$ on every value, and then projecting these costs from $w_1$ to $w_\varnothing$ by enforcing NC. However, if $w_\varnothing = w_1 = w_2 = 0$ and $k$ is strictly greater than 11, BAC remains idle here. We can however simply improve BAC by directly taking into account the minimum possible cost of the cost function $w_{12}$ over all possible assignments given the current domains.

**Definition 4.1** *A cost function $w_S$ is $\varnothing$-inverse consistent ($\varnothing$-IC) iff:*

$$\exists t_S \in \ell_S, w_S(t_S) = 0$$

*Such a tuple $t_S$ is called a support for $w_S$. A WCN is $\varnothing$-IC iff every cost function (except $w_\varnothing$) is $\varnothing$-IC.*

Enforcing $\varnothing$-IC can always be done as follows: for every cost function $w_S$ with a non empty scope, the minimum cost assignment of $w_S$ given the current variable domains is computed. The cost $\alpha$ of this assignment is then subtracted from all the tuple costs in $w_S$ and added to $w_\varnothing$. This creates at least one support in $w_S$ and makes the cost function $\varnothing$-IC. For a given cost function $w_S$, this is done by the Project() function of Algorithm 3.

In order to strengthen BAC, a natural idea is to combine it with $\varnothing$-IC. We will call BAC$^\varnothing$ the resulting combination of BAC and $\varnothing$-IC. To enforce BAC$^\varnothing$, the previous algorithm is modified by first adding a call to the Project() function (see line 53 of Algorithm 3). Moreover, to maintain BAC whenever $w_\varnothing$ is modified by projection, every variable is tested for possible pruning at line 66 and put back in $Q$ in case of domain change. Note that the subtraction applied to all constraint tuples at line 75 can be done in constant time without modifying the constraint by using an additional $\Delta^{w_S}$ data-structure, similar to the $\Delta$ data-structure introduced by Cooper and Schiex (2004). This data-structure keeps track of the cost which has been projected from $w_S$ to $w_\varnothing$. This feature makes it possible to leave the original costs unchanged during the enforcement of the local consistency. For example, for any $t_S \in \ell_S$, $w_S(t)$ refers to $\mathbf{w_S}(t) \ominus \Delta^{w_S}$, where $\mathbf{w_S}(t)$ denotes the original cost. Note that $\Delta^{w_S}$, which will be later used in a confluence proof, precisely contains the amount of cost which has been moved from $w_S$ to $w_\varnothing$. The whole algorithm is described in Algorithm 3. We highlighted in black the parts which are different from Algorithm 2 whereas the unchanged parts are in gray.

**Proposition 4.2 (Time and space complexity)** *For a WCN with maximum arity $r$ of the constraints, enforcing BAC$^\varnothing$ with Algorithm 3 can be enforced in $O(n^2 r^2 d^{r+1})$ time using $O(n + er)$ memory space.*

**Proof:** Every variable is pushed at most $O(d)$ times in $Q$, thus the **foreach** at line 51 (resp. line 55) loops at most $O(erd)$ (resp. $O(er^2d)$) times. The projection on line 53 takes $O(d^r)$ time. The operation at line 57 can be carried out in $O(d^{r-1})$ time. The overall time spent inside the **if** of the PruneInf() function is bounded by $O(ed)$. Thus the overall time spent in the loop at line 51 (resp. line 55) is bounded by $O(er^2d^{r+1})$ (resp. $O(er^2d^r)$).

The flag on line 66 is true when $w_\varnothing$ increases, and so it cannot be true more than $k$ times (assuming integer costs). If the flag is true, then we spend $O(n)$ time to check all the bounds of the variables. Thus, the time complexity under the **if** is bounded by $O(\min\{k, nd\} \times n)$. To sum up, the overall time complexity is $O(er^2d^{r+1} + \min\{k, nd\} \times n)$, which is bounded by $O(n^2 r^2 d^{r+1})$.

The space complexity is given by the $\Delta$, $w^{\inf}$, $w^{\sup}$ and $\Delta^{w_S}$ data-structures which sums up to $O(n + re)$ for a WCN with an arity bounded by $r$. $\square$

The time complexity of the algorithm enforcing BAC$^\varnothing$ is multiplied by $d$ compared to BAC without $\varnothing$-IC. This is a usual trade-off between the strength of a local property and the time spent to enforce it. However, the space complexity is still independent of $d$. Moreover, like BAC, BAC$^\varnothing$ is confluent.

**Proposition 4.3 (Confluence)** *Enforcing BAC$^\varnothing$ on a given problem always leads to a unique WCN.*

**Proof:** The proof is similar to the proof of Proposition 3.3. However, because of the possible cost movements induced by projections, BAC$^\varnothing$ transforms the original problem $\mathcal{P}$ in more complex ways, allowing either pruning domains (BAC) or moving costs from cost

---

**Algorithm 3**: Algorithm enforcing $\text{BAC}^\varnothing$

---

**44 Procedure** $\text{BAC}^\varnothing(\mathcal{X}, \mathcal{D}, \mathcal{W}, k)$

**45**    $Q \leftarrow \mathcal{X}$ ;

**46**    $w^{\inf}(\cdot) \leftarrow 0$ ; $w^{\sup}(\cdot) \leftarrow 0$ ; $\Delta^{\inf}(\cdot, \cdot) \leftarrow 0$ ; $\Delta^{\sup}(\cdot, \cdot) \leftarrow 0$ ;

**47**    **while** $(Q \neq \varnothing)$ **do**

**48**      $x_j \leftarrow pop(Q)$ ;

**49**      $flag \leftarrow$ false ;

**51**      **foreach** $w_S \in \mathcal{W}, x_j \in S$ **do**

**53**        **if** $\text{Project}(w_S)$ **then** $flag \leftarrow$ true ;

**55**        **foreach** $x_i \in S$ **do**

**57**          $\alpha \leftarrow \min_{t_S \in \ell_S, \inf(x_i) \in t_S} w_S(t_S)$ ;

**58**          $w^{\inf}(x_i) \leftarrow w^{\inf}(x_i) \ominus \Delta^{\inf}(x_i, w_S) \oplus \alpha$ ;

**59**          $\Delta^{\inf}(x_i, w_S) \leftarrow \alpha$ ;

**60**          **if** $\text{PruneInf}(x_i)$ **then** $Q \leftarrow Q \cup \{x_i\}$ ;

**61**          $\alpha \leftarrow \min_{t_S \in \ell_S, \sup(x_i) \in t_S} w_S(t_S)$ ;

**62**          $w^{\sup}(x_i) \leftarrow w^{\sup}(x_i) \ominus \Delta^{\sup}(x_i, w_S) \oplus \alpha$ ;

**63**          $\Delta^{\sup}(x_i, w_S) \leftarrow \alpha$ ;

**64**          **if** $\text{PruneSup}(x_i)$ **then** $Q \leftarrow Q \cup \{x_i\}$ ;

**66**      **if** $(flag)$ **then**

**67**        **foreach** $x_i \in \mathcal{X}$ **do**

**68**          **if** $\text{PruneInf}(x_i)$ **then** $Q \leftarrow Q \cup \{x_i\}$ ;

**69**          **if** $\text{PruneSup}(x_i)$ **then** $Q \leftarrow Q \cup \{x_i\}$ ;

**70 Function** $\text{Project}(w_S)$ *: boolean*

**71**    $\alpha \leftarrow \min_{t_S \in \ell_S} w_S(t_S)$ ;

**72**    **if** $(\alpha > 0)$ **then**

**73**      $w_\varnothing \leftarrow w_\varnothing \oplus \alpha$ ;

**75**      $w_S(\cdot) \leftarrow w_S(\cdot) \ominus \alpha$ ;

**76**      **return** true;

**77**    **else** **return** false;

---

functions to $w_\varnothing$. The set of problems that will be considered needs therefore to take this into account. Instead of being just defined by its domains, a WCN reached by $\text{BAC}^\varnothing$ is also characterized by the amount of cost that has been moved from each cost function $w_S$ to $w_\varnothing$. This quantity is already denoted by $\Delta^{w_S}$ in Section 4, on page 603. We therefore consider the set $\wp_2(\mathcal{P})$ defined by:

$$\left\{ (\langle \mathcal{X}, \mathcal{D}', \mathcal{W}, k \rangle, \{\Delta^w : w \in \mathcal{W}\}) : \forall i \in [1, n], D'(x_i) \subseteq D(x_i), \forall w \in \mathcal{W}, \Delta^w \in [0, k] \right\}$$

We can now define the relation $\sqsubseteq$ on $\wp_2(\mathcal{P})$:

$$\mathcal{P}_1 \sqsubseteq \mathcal{P}_2 \Leftrightarrow ((\forall w \in \mathcal{W}, \Delta_1^w \geq \Delta_2^w) \wedge (\forall x_i \in \mathcal{X}, D_1(x_i) \subseteq D_2(x_i)))$$

This relation is reflexive, transitive and antisymmetric. The first two properties can be easily verified. Suppose now that $(\mathcal{P}_1, \mathcal{P}_2) \in \wp_2^2(\mathcal{P})$ and that $(\mathcal{P}_1 \sqsubseteq \mathcal{P}_2) \wedge (\mathcal{P}_2 \sqsubseteq \mathcal{P}_1)$. We have thus $(\forall w \in \mathcal{W}, \Delta_w = \Delta'_w) \wedge (\forall x_i \in \mathcal{X}, D(x_i) = D'(x_i))$. This ensures that the domains, as well as the amounts of cost projected by each cost function, are the same. Thus, the problems are the same and $\sqsubseteq$ is antisymmetric.

Besides, $\langle \wp_2(\mathcal{P}), \sqsubseteq \rangle$ is a complete lattice, since:

$$\forall (\mathcal{P}_1, \mathcal{P}_2) \in \wp_2^2(\mathcal{P}),$$
$$\text{glb}(\mathcal{P}_1, \mathcal{P}_2) = (\langle \mathcal{X}, \{D_1(x_i) \cap D_2(x_i) : i \in [1, n]\}, \mathcal{W}, k \rangle, \{\max\{\Delta_1^w, \Delta_2^w\} : w \in \mathcal{W}\})$$
$$\text{lub}(\mathcal{P}_1, \mathcal{P}_2) = (\langle \mathcal{X}, \{D_1(x_i) \cup D_2(x_i) : i \in [1, n]\}, \mathcal{W}, k \rangle, \{\min\{\Delta_1^w, \Delta_2^w\} : w \in \mathcal{W}\})$$

and both of them are in $\wp_2(\mathcal{P})$.

Every enforcement of $\text{BAC}^\varnothing$ follows from the application of functions from a set of functions $F_{\text{BAC}^\varnothing}$ which may remove the maximum or minimum domain bound (same definition as for BAC) or may project cost from cost functions to $w_\varnothing$. For a given cost function $w \in \mathcal{W}$, such a function keeps the instance unchanged if the minimum $\alpha$ of $w$ is 0 over possible tuples. Otherwise, if $\alpha > 0$, the problem returned is derived from $\mathcal{P}$ by projecting an amount of cost $\alpha$ from $w$ to $w_\varnothing$. These functions are easily shown to be order preserving for $\sqsubseteq$.

As in the proof of Proposition 3.3, we can define the lattice $\wp_2^\star(\mathcal{P})$, which is the intersection of the sets of fixed points of the functions $f \in F_{\text{BAC}^\varnothing}$. $\wp_2^\star(\mathcal{P})$ is not empty, since $(\langle \mathcal{X}, \{\varnothing, \ldots, \varnothing\}, \mathcal{W}, k \rangle, \{k, \ldots, k\})$ is in it. As in the proof of proposition 3.3, and since Algorithm 3 terminates, we can conclude that this algorithm is confluent, and that it results in $\text{lub}(\wp_2^\star(\mathcal{P}))$. □

## 5. Exploiting Cost Function Semantics in $\text{BAC}^\varnothing$

In crisp AC, several classes of binary constraints make it possible to enforce AC significantly faster (in $O(ed)$ instead of $O(ed^2)$, as shown by Van Hentenryck et al., 1992). Similarly, it is possible to exploit the semantics of the cost functions to improve the time complexity of $\text{BAC}^\varnothing$ enforcement. As the proof of Proposition 4.2 shows, the dominating factors in this complexity comes from the complexity of computing the minimum of cost functions during projection at lines 53 and 57 of Algorithm 3. Therefore, any cost function property that makes these computations less costly may lead to an improvement of the overall time complexity.

**Proposition 5.1** *In a binary WCN, if for any cost function $w_{ij} \in \mathcal{W}$ and for any sub-intervals $E_i \subseteq D(x_i)$, $E_j \subseteq D(x_j)$, the minimum of $w_{ij}$ over $E_i \times E_j$ can be found in time $\mathcal{O}(d)$, then the time complexity of enforcing $BAC^\varnothing$ is $\mathcal{O}(n^2 d^2)$.*

**Proof:** This follows directly from the proof of Proposition 4.2. In this case, the complexity of projection at line 53 is only in $\mathcal{O}(d)$ instead of $\mathcal{O}(d^2)$. Thus the overall time spent in the loop at line 51 is bounded by $\mathcal{O}(ed^2)$ and the overall complexity is $\mathcal{O}(ed^2 + n^2 d) \leq \mathcal{O}(n^2 d^2)$.
$\square$

**Proposition 5.2** *In a binary WCN, if for any cost function $w_{ij} \in \mathcal{W}$ and for any sub-intervals $E_i \subseteq D(x_i)$, $E_j \subseteq D(x_j)$, the minimum of $w_{ij}$ over $E_i \times E_j$ can be found in constant time, then the time complexity of enforcing $BAC^\varnothing$ is $\mathcal{O}(n^2 d)$.*

**Proof:** This follows again from the proof of Proposition 4.2. In this case, the complexity of projection at line 53 is only in $\mathcal{O}(1)$ instead of $\mathcal{O}(d^2)$. Moreover, the operation at line 57 can be carried out in time $\mathcal{O}(1)$ instead of $\mathcal{O}(d)$. Thus, the overall time spent in the loop at line 51 is bounded by $O(ed)$ and the overall complexity is $\mathcal{O}(ed + n^2 d) = \mathcal{O}(n^2 d)$. $\square$

These two properties are quite straightforward and one may wonder if they have non trivial usage. They can actually be directly exploited to generalize the results presented by Van Hentenryck et al. (1992) for functional, anti-functional and monotonic constraints. In the following sections, we show that functional, anti-functional and semi-convex cost functions (which include monotonic cost functions) can indeed benefit from an $\mathcal{O}(d)$ speedup factor by application of Proposition 5.1. For monotonic cost functions and more generally any convex cost function, a stronger speedup factor of $\mathcal{O}(d^2)$ can be obtained by Proposition 5.2.

## 5.1 Functional Cost Functions

The notion of functional constraint can be extended to cost functions as follows:

**Definition 5.3** *A cost function $w_{ij}$ is functional w.r.t. $x_i$ iff:*

- $\forall (v_i, v_j) \in D(x_i) \times D(x_j), w_{ij}(v_i, v_j) \in \{0, \alpha\}$ *with $\alpha \in [1, k]$*

- $\forall v_i \in D(x_i)$, *there is at most one value $v_j \in D(x_j)$ such that $w_{ij}(v_i, v_j) = 0$. When it exists, this value is called the functional support of $v_i$.*

We assume in the rest of the paper that the functional support can be computed in constant time. For example, the cost function $w_{ij}^= = \begin{cases} 0 & \text{if } x_i = x_j \\ 1 & \text{otherwise} \end{cases}$ is functional. In this case, the functional support of $v_i$ is itself. Note that for $k = 1$, functional cost functions represent functional constraints.

**Proposition 5.4** *The minimum of a functional cost function $w_{ij}$ w.r.t. $x_i$ can always be found in $O(d)$.*

**Proof:** For every value $v_i$ of $x_i$, one can just check if the functional support of $v_i$ belongs to the domain of $x_j$. This requires $\mathcal{O}(d)$ checks. If this is never the case, then the minimum of the cost function is known to be $\alpha$. Otherwise, it is 0. The result follows. □

## 5.2 Anti-Functional and Semi-Convex Cost Functions

**Definition 5.5** *A cost function $w_{ij}$ is anti-functional w.r.t. the variable $x_i$ iff:*

- $\forall (v_i, v_j) \in D(x_i) \times D(x_j), w_{ij}(v_i, v_j) \in \{0, \alpha\}$ *with $\alpha \in [1, k]$*

- $\forall v_i \in D(x_i)$, *there is at most one value $v_j \in D(x_j)$ such that $w_{ij}(v_i, v_j) = \alpha$. When it exists, this value is called the anti-support of $v_i$.*

The cost function $w_{ij}^{\neq} = \begin{cases} 0 & \text{if } x_i \neq x_j \\ 1 & \text{otherwise} \end{cases}$ is an example of an anti-functional cost function. In this case, the anti-support of $v_i$ is itself. Note that for $k = 1$, anti-functional cost functions represent anti-functional constraints.

Anti-functional cost functions are actually a specific case of semi-convex cost functions, a class of cost functions that appear for example in temporal constraint networks with preferences (Khatib, Morris, Morris, & Rossi, 2001).

**Definition 5.6** *Assume that the domain $D(x_j)$ is contained in a set $D_j$ totally ordered by the order $<^j$.*

*A function $w_{ij}$ is semi-convex w.r.t. $x_i$ iff $\forall \beta \in [0, k], \forall v_i \in D_i$, the set $\{v_j \in D_j : w_{ij}(v_i, v_j) \geq \beta\}$, called the $\beta$-support of $v_i$, defines an interval over $D_j$ according to $<^j$.*

Semi-convexity relies on the definition of intervals defined in a totally ordered discrete set denoted $D_j$, and ordered by $<^j$. Even if they may be identical, it is important to avoid confusion between the order $\prec_j$ over $D(x_j)$, used to define interval domains for bounds arc consistency, and the order $<^j$ over $D_j$ used to define intervals for semi-convexity. In order to guarantee constant time access to the minimum and maximum elements of $D(x_j)$ according to $<^j$ (called the $<^j$-bounds of the domain), we assume that $<^j = \prec_j$ or $<^j = \succ_j$[1]. In this case, the $<^j$-bounds and the domain bounds are identical.

One can simply check that anti-functional cost functions are indeed semi-convex: in this case, the $\beta$-support of any value is either the whole domain ($\beta = 0$), reduced to one point ($0 < \beta \leq \alpha$) or to the empty set (otherwise). Another example is the cost function $w_{ij} = x_i^2 - x_j^2$ which is semi-convex w.r.t. $x_i$.

**Proposition 5.7** *The minimum of a cost function $w_{ij}$ which is semi-convex w.r.t. one of its variables can always be found in $O(d)$.*

**Proof:** We will first show that, if $w_{ij}$ is semi-convex w.r.t. to one of its variables (let say $x_i$), then for any value $v_i$ of $x_i$, the cost function $w_{ij}$ must be minimum at one of the $<^j$-bounds of $D_j$.

---

1. This restriction could be removed using for example a doubly-linked list data-structure over the values in $D(x_j)$, keeping the domain sorted according to $<^j$ and allowing constant time access and deletion but this would be at the cost of linear space which we cannot afford in the context of BAC.

Assume $x_i$ is set to $v_i$. Let $\beta_b$ be the lowest cost reached on either of the two $<^j$-bounds of the domain. Since $w_{ij}$ is semi-convex, then $\{v_j \in D_j : w_{ij}(v_i, v_j) \geq \beta_b\}$ is an interval, and thus every cost $w_{ij}(v_i, v_j)$ is not less than $\beta_b$ for every value of $D_j$. Therefore, at least one of the two $<^j$-bounds has a minimum cost.

In order to find the global minimum of $w_{ij}$, we can restrict ourselves to the $<^j$-bounds of the domain of $x_j$ for every value of $x_i$. Therefore, only $2d$ costs need to be checked. $\square$

From Proposition 5.1, we can conclude

**Corollary 5.8** *In a binary WCN, if all cost functions are functional, anti-functional or semi-convex, the time complexity of enforcing $BAC^\varnothing$ is $\mathcal{O}(n^2 d^2)$ only.*

### 5.3 Monotonic and Convex Cost Functions

**Definition 5.9** *Assume that the domain $D(x_i)$ (resp. $D(x_j)$) is contained in a set $D_i$ (resp. $D_j$) totally ordered by the order $<^i$ (resp. $<^j$).*

*A cost function $w_{ij}$ is* monotonic *iff:*

$$\forall (v_i, v_i', v_j, v_j') \in D_i^2 \times D_j^2, v_i' \leq^i v_i \wedge v_j' \geq^j v_j \Rightarrow w_{ij}(v_i', v_j') \leq w_{ij}(v_i, v_j)$$

The cost function $w_{ij}^{\leq} = \begin{cases} 0 & \text{if } x_i \leq x_j \\ 1 & \text{otherwise} \end{cases}$ is an example of a monotonic cost function. Monotonic cost functions are actually instances of a larger class of functions called convex functions.

**Definition 5.10** *A function $w_{ij}$ is* convex *iff it is semi-convex w.r.t. each of its variables.*

For example, $w_{ij} = x_i + x_j$ is convex.

**Proposition 5.11** *The minimum of a convex cost function can always be found in constant time.*

**Proof:** Since the cost function is semi-convex w.r.t. each of its variable, we know from the proof of Proposition 5.7 that it must reach a minimum cost on one of the $<^j$-bounds of the domain of $x_j$ and similarly for $x_i$. There are therefore only four costs to check in order to compute the minimum cost. $\square$

From Proposition 5.2, we conclude that

**Corollary 5.12** *In a binary WCN, if all cost functions are convex, then the time complexity of enforcing $BAC^\varnothing$ is $\mathcal{O}(n^2 d)$ only.*

One interesting example for a convex cost function is $w_{ij} = \max\{x_i - x_j + cst, 0\}$. This type of cost function, which can be efficiently filtered by $BAC^\varnothing$, may occur in temporal reasoning problems and is also used in our RNA gene localization problem for specifying preferred distances between elements of a gene.

## 6. Comparison with Crisp Bounds Consistency

Petit et al. (2000) have proposed to transform WCNs into crisp constraint networks with extra cost variables. In this transformation, every cost function is *reified* into a constraint, which applies on the original cost function scope augmented by one extra variable representing the assignment cost. This reification of costs into domain variables transforms a WCN in a crisp CN with more variables and augmented arities. As proposed by Petit et al., it can be achieved using meta-constraints, i.e. logical operators applied to constraints. Given this relation between WCNs and crisp CNs and the relation between $\text{BAC}^\varnothing$ and bounds consistency, it is natural to wonder how $\text{BAC}^\varnothing$ enforcing relates to just enforcing bounds consistency on the reified version of a WCN.

In this section we show that $\text{BAC}^\varnothing$ is in some precise sense stronger than enforcing bounds consistency on the reified form. This is a natural consequence of the fact that the domain filtering in BAC is based on the combined cost of several cost functions instead of taking each constraint separately in bounds consistency. We first define the reification process precisely. We then show that $\text{BAC}^\varnothing$ can be stronger than the reified bounds consistency on one example and conclude by proving that it can never be weaker.

The following example introduces the cost reification process.

**Example 6.1** Consider the WCN in Figure 2(a). It contains two variables $x_1$ and $x_2$, one binary cost function $w_{12}$, and two unary cost functions $w_1$ and $w_2$. For the sake of clarity, every variable or constraint in the reified hard model, described on Figure 2(b), will be indexed by the letter R.

First of all, we model every cost function by a hard constraint, and express that assigning $b$ to $x_1$ yields a cost of 1. We create a new variable $x_{1R}^C$, the *cost variable of $w_1$*, that stores the cost of any assignment of $x_1$. Then, we replace the unary cost function $w_1$ by a binary constraint $c_{1R}$ that involves $x_1$ and $x_{1R}^C$, such that if a value $v_1$ is assigned to $x_1$, then $x_{1R}^C$ should take the value $w_1(v_1)$. We do the same for the unary cost function $w_2$. The idea is the same for the binary cost function $w_{12}$: we create a new variable $x_{12R}^C$, and we replace $w_{12}$ by a ternary constraint $c_{12R}$, that makes sure that for any assignment of $x_1$ and $x_2$ to $v_1$ and $v_2$ respectively, $x_{12R}^C$ takes the value $w_{12}(v_1, v_2)$. Finally, a global cost constraint $c_R^C$ that states that the sum of the cost variables should be less than $k$ is added: $x_{1R}^C + x_{2R}^C + x_{12R}^C < k$. This completes the description of the reified cost hard constraint network.

We can now define more formally the reification process of a WCN.

**Definition 6.2** *Consider the WCN $\mathcal{P} = \langle \mathcal{X}, \mathcal{D}, \mathcal{W}, k \rangle$. Let $\text{reify}(\mathcal{P}) = \langle \mathcal{X}_R, \mathcal{D}_R, \mathcal{W}_R \rangle$ be the crisp CN such that:*

- *the set $\mathcal{X}_R$ contains one variable $x_{iR}$ for every variable $x_i \in \mathcal{X}$, augmented with an extra cost variable $x_{S_R}^C$ per cost function $w_S \in \mathcal{W} - \{w_\varnothing\}$.*

- *the domains $\mathcal{D}_R$ are:*

  - *$D_R(x_{iR}) = D(x_i)$ for the $x_{iR}$ variables, with domain bounds $lb_{iR}$ and $ub_{iR}$,*
  - *$[lbs_{S_R}^C, ubs_{S_R}^C] = [0, k-1]$ for the $x_{S_R}^C$ variables.*

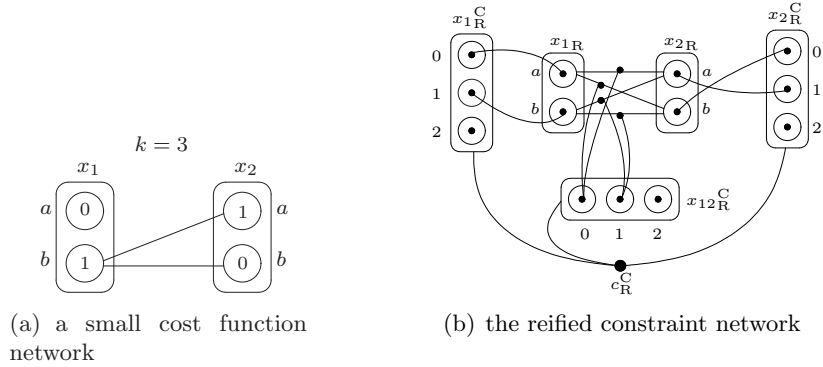

(a) a small cost function network

(b) the reified constraint network

Figure 2: A small cost function network and its reified counterpart.

- *the set $\mathcal{W}_R$ of constraints contains:*

  - *$c_{SR} = \{(t, w_S(t)) : t \in \ell_S, w_\varnothing \oplus w_S(t) < k\}$, with scope $S \cup \{x_{SR}^C\}$, for every cost function $w_S \in \mathcal{W}$,*

  - *$c_R^C$ is defined as $(w_\varnothing \oplus \sum_{w_S \in \mathcal{W}} x_{SR}^C < k)$, an extra constraint that makes sure that the sum of the cost variables is strictly less than $k$.*

It is simple to check that the problem reify($\mathcal{P}$) has a solution iff $\mathcal{P}$ has a solution and the sum of the cost variables in a solution is the cost of the corresponding solution (defined by the values of the $x_{iR}$ variables) in the original WCN.

**Definition 6.3** *Let $\mathcal{P}$ be a problem, $\ell$ and $\ell'$ two local consistency properties. Let $\ell(\mathcal{P})$ be the problem obtained after filtering $\mathcal{P}$ by $\ell$. $\ell$ is said to be* not weaker *than $\ell'$ iff $\ell'(\mathcal{P})$ emptiness implies $\ell(\mathcal{P})$ emptiness.*

*$\ell$ is said to be* stronger *than $\ell'$ iff it is not weaker than $\ell'$, and if there exists a problem $\mathcal{P}$ such that $\ell'(\mathcal{P})$ is not empty but $\ell(\mathcal{P})$ is empty.*

This definition is practically very significant since the emptiness of a filtered problem is the event that generates backtracking in tree search algorithms used for solving CSP and WCSP.

**Example 6.4** Consider the WCN defined by three variables ($x_1$, $x_2$ and $x_3$) and two binary cost functions ($w_{12}$ and $w_{13}$). $D(x_1) = \{a, b, c, d\}$, $D(x_2) = D(x_3) = \{a, b, c\}$ (we assume that $a \prec b \prec c \prec d$). The costs of the binary cost functions are described in Figure 3. Assume that $k = 2$ and $w_\varnothing = 0$.

One can check that the associated reified problem is already bounds consistent and obviously not empty. For example, a support of the minimum bound of the domain of $x_{1R}$ w.r.t. $c_{12R}$ is $(a, a, 1)$, a support of its maximum bound is $(d, a, 1)$. Supports of the maximum and minimum bounds of the domain of $x_{12R}^C$ w.r.t. $c_{12R}$ are $(b, a, 0)$ and $(a, a, 1)$ respectively. Similarly, one can check that all other variable bounds are also supported on all the constraints that involve them.

$$
\begin{array}{cc}
\begin{array}{c}
(x_1) \\
\begin{array}{c}
\phantom{a} \\
(x_2) \\
\phantom{a}
\end{array}
\begin{array}{c}
\begin{array}{cccc}
a & b & c & d
\end{array} \\
\begin{array}{c}
a \\
b \\
c
\end{array}
\left[\begin{array}{cccc}
1 & 0 & 2 & 1 \\
1 & 0 & 2 & 1 \\
1 & 0 & 2 & 1
\end{array}\right]
\end{array}
\end{array}
&
\begin{array}{c}
(x_1) \\
\begin{array}{c}
\phantom{a} \\
(x_3) \\
\phantom{a}
\end{array}
\begin{array}{c}
\begin{array}{cccc}
a & b & c & d
\end{array} \\
\begin{array}{c}
a \\
b \\
c
\end{array}
\left[\begin{array}{cccc}
1 & 2 & 0 & 1 \\
1 & 2 & 0 & 1 \\
1 & 2 & 0 & 1
\end{array}\right]
\end{array}
\end{array}
\end{array}
$$

Figure 3: Two cost matrices.

However, the original problem is not BAC since for example, the value $a$, the minimum bound of the domain of $x_1$, does not satisfy the BAC property:

$$
w_\varnothing \oplus \sum_{w_S \in \mathcal{W}, x_1 \in S} \left\{ \min_{t_S \in \ell_S, a \in t_S} w_S(t_S) \right\} < k
$$

This means that the value $a$ can be deleted by BAC filtering. By symmetry, the same applies to the maximum bound of $x_1$ and ultimately, the problem inconsistency will be proved by BAC. This shows that bounds consistency on the reified problem cannot be stronger than BAC on the original problem.

We will now show that $\mathrm{BAC}^\varnothing$ is actually stronger than bounds consistency applied on the reified WCN. Because $\mathrm{BAC}^\varnothing$ consistency implies non-emptiness (since it requires the existence of assignments of cost 0 in every cost function) we will start from any $\mathrm{BAC}^\varnothing$ consistent WCN $\mathcal{P}$ (therefore not empty) and prove that filtering the reified problem $\mathrm{reify}(\mathcal{P})$ by bounds consistency does not lead to an empty problem.

**Lemma 6.5** *Let $\mathcal{P}$ be a $BAC^\varnothing$ consistent binary WCN. Then filtering $\mathrm{reify}(\mathcal{P})$ by bounds consistency does not produce an empty problem.*

**Proof:** We will prove here that bounds consistency will just reduce the maximum bounds of the domains of the cost variables $x_{S\mathrm{R}}^{\mathrm{C}}$ to a non empty set and leave all other domains unchanged.

More precisely, the final domain of $x_{S\mathrm{R}}^{\mathrm{C}}$ will become $[0, \max\{w_S(t) : t \in \ell_S, w_\varnothing \oplus w_S(t) < k\}]$. Note that this interval is not empty because the network is $\mathrm{BAC}^\varnothing$ consistent which means that every cost function has an assignment of cost 0 (by $\varnothing$-IC) and $w_\varnothing < k$ (or else the bounds of the domains could not have supports and the problem would not be BAC).

To prove that bounds consistency will not reduce the problem by more than this, we simply prove that the problem defined by these domain reductions only is actually bounds consistent.

All the bounds consistency required properties apply to the bounds of the domains of the variables of $\mathrm{reify}(\mathcal{P})$. Let us consider every type of variable in this reified reduced problem:

- reified variables $x_{i\mathrm{R}}$. Without loss of generality, assume that the minimum bound $lb_{i\mathrm{R}}$ of $x_{i\mathrm{R}}$ is not bounds consistent (the symmetrical reasoning applies to the maximum bound). This means it would have no support with respect to a given reified constraint

$c_{S\mathrm{R}}, x_i \in S$. However, by BAC, we have

$$w_\varnothing \oplus \min_{t \in \ell_S, lb_{i\mathrm{R}} \in t} w_S(t) < k$$

and so $\quad \exists t \in \ell_S, lb_{i\mathrm{R}} \in t, \quad w_S(t) \leq \max\{w_S(t) : t \in \ell_S, w_\varnothing \oplus w_S(t) < k\}$

which means that $lb_{i\mathrm{R}}$ is supported w.r.t. $c_{S\mathrm{R}}$.

- cost variables. The minimum bound of all cost variables are always bounds consistent w.r.t. the global constraint $c_\mathrm{R}^\mathrm{C}$ because this constraint is a "less than" inequality. Moreover, since the minimum bounds of the cost variables are set to 0, they are also consistent w.r.t. the reified constraints, by the definition of $\varnothing$-inverse consistency.

  Consider the maximum bound $ubs_{S\mathrm{R}}^\mathrm{C}$ of a cost variable in the reduced reified problem. Remember it is defined as $\max\{w_S(t) : t \in \ell_S, w_\varnothing \oplus w_S(t) < k\}$, and so $w_\varnothing \oplus ubs_{S\mathrm{R}}^\mathrm{C} < k$. The minimum bounds of all other cost variables in the reified problem, which are 0, form a support of $ubs_{S\mathrm{R}}^\mathrm{C}$ w.r.t. the global constraint $c_\mathrm{R}^\mathrm{C}$. So $ubs_{S\mathrm{R}}^\mathrm{C}$ cannot be removed by bounds consistency.

$\qquad\qquad\qquad\qquad\qquad\qquad\qquad\qquad\qquad\qquad\qquad\qquad\qquad\qquad\qquad\qquad\quad\square$

We will now prove the final assertion:

**Proposition 6.6** *$BAC^\varnothing$ is stronger than bounds consistency.*

**Proof:** Lemma 6.5 shows that $BAC^\varnothing$ is not weaker than bounds consistency. Then, example 6.4 is an instance where BAC, and therefore $BAC^\varnothing$ is actually stronger than bounds consistency after reification. $\qquad\qquad\qquad\qquad\qquad\qquad\qquad\qquad\qquad\qquad\quad\square$

A filtering related to $BAC^\varnothing$ could be achieved in the reified approach by an extra shaving process where each variable is assigned to one of its domain bounds and this bound is deleted if an inconsistency is found after enforcing bounds consistency (Lhomme, 1993).

## 7. Other Related Works

The Definition 3.1 of BAC is closely related to the notion of arc consistency counts introduced by Freuder and Wallace (1992) for Max-CSP processing. The Max-CSP can be seen as a very simplified form of WCN where cost functions only generate costs of 0 or 1 (when the associated constraint is violated). Our definition of BAC can be seen as an extension of AC counts allowing dealing with arbitrary cost functions, including the usage of $w_\varnothing$ and $k$, and applied only to domain bounds as in bounds consistency. The addition of $\varnothing$-IC makes $BAC^\varnothing$ more powerful.

Dealing with large domains in Max-CSP has also been considered in the Range-Based Algorithm, again designed for Max-CSP by Petit, Régin, and Bessière (2002). This algorithm uses reversible directed arc consistency (DAC) counts and exploits the fact that in Max-CSP, several successive values in a domain may have the same DAC counts. The algorithm intimately relies on the fact that the problem is a Max-CSP problem, defined by a set of constraints and actively uses bounds consistency dedicated propagators for the constraints in the Max-CSP. In this case the number of different values reachable by the DAC counters of a variable is bounded by the degree of the variable, which can be much smaller than the domain size. Handling intervals of values with a same DAC cost as one value allows space and time savings. For arbitrary binary cost functions, the translation into constraints could generate up to $d^2$ constraints for a single cost function and makes the scheme totally impractical.

Several alternative definition of bounds consistency exist in crisp CSPs (Choi et al., 2006). Our extension to WCSP is based on $bounds(\mathcal{D})$ or $bounds(\mathcal{Z})$ consistencies (which are equivalent on intervals). For numerical domains, another possible weaker definition of bounds consistency is $bounds(\mathcal{R})$ consistency, which is obtained by a relaxation to real numbers. It has been shown by Choi et al. that $bounds(\mathcal{R})$ consistency can be checked in polynomial time on some constraints whereas $bounds(\mathcal{D})$ or $bounds(\mathcal{Z})$ is NP-hard (eg. for linear equality). The use of this relaxed version in the WCSP context together with intentional description of cost functions would have the side effect of extending the cost domain from integer to real numbers. Because extensional or algorithmical description of integer cost functions is more general and frequent in our problems, this possibility was not considered. Since cost comparison is the fundamental mechanism used for pruning in WCSP, a shift to real numbers for costs would require a safe floating number implementation both in the local consistency enforcing algorithms and in the branch and bound algorithm.

## 8. Experimental Results

We experimented bounds arc consistency on two benchmarks translated into weighted CSPs. The first benchmark is from AI planning and scheduling. It is a mission management benchmark for agile satellites (Verfaillie & Lemaître, 2001; de Givry & Jeannin, 2006). The maximum domain size of the temporal variables is 201. This reasonable size and the fact that there are only binary cost functions allows us to compare $BAC^\varnothing$ with strong local consistencies such as EDAC*. Additionally, this benchmark has also been modeled using the reified version of WCN, thus allowing for an experimental counterpart of the theoretical comparison of Section 6.

The second benchmark comes from bioinformatics and models the problem of the localization of non-coding RNA molecules in genomes (Thébault et al., 2006; Zytnicki et al., 2008). Our aim here is mostly to confirm that bounds arc consistency is useful and practical on a real complex problem with huge domains, which can reach several millions.

### 8.1 A Mission Management Benchmark for Agile Satellites

We solved a simplified version described by de Givry and Jeannin (2006) of a problem of selecting and scheduling earth observations for agile satellites. A complete description of the problem is given by Verfaillie and Lemaître (2001). The satellite has a pool of candidate photographs to take. It must select and schedule a subset of them on each pass above a certain strip of territory. The satellite can only take one photograph at a time (disjunctive scheduling). A photograph can only be taken during a time window that depends on the location photographed. Minimal repositioning times are required between two consecutive photographs. All physical constraints (time windows and repositioning times) must be met, and the sum of the revenues of the selected photographs must be maximized. This is equivalent to minimizing the "rejected revenues" of the non selected photographs.

Let $N$ be the number of candidate photographs. We define $N$ decision variables representing the acquisition starting times of the candidate photographs. The domain of each variable is defined by the time window of its corresponding photograph plus an extra domain value which represents the fact that the photograph is not selected. As proposed by de Givry and Jeannin (2006), we create a binary hard constraint for every pair of photographs (resulting in a complete constraint graph) which enforces the minimal repositioning times if both photographs are selected (represented by a disjunctive constraint). For each photograph, a unary cost function associates its rejected revenue to the corresponding extra value.

In order to have a better filtering, we moved costs from unary cost functions inside the binary hard constraints in a preprocessing step. This allows bounds arc consistency filtering to exploit the revenue information and the repositioning times jointly, possibly increasing $w_\varnothing$ and the starting times of some photographs. To achieve this, for each variable $x_i$, the unary cost function $w_i$ is successively combined (using $\oplus$) with each binary hard constraint $w_{ij}$ that involves $x_i$. This yields $N-1$ new binary cost functions $w'_{ij}$ defined as $w'_{ij}(t) = w_{ij}(t) \oplus w_i(t[x_i])$, having both hard $(+\infty)$ and soft weights. These binary cost functions $w'_{ij}$ replace the unary cost function $w_i$ and the $N-1$ original binary hard constraints $w_{ij}$. Notice that this transformation has the side effect of multiplying all soft weights by $N-1$. This does preserve the equivalence with the original problem since all finite weights are just multiplied by the same constant $(N-1)$.

The search procedure is an exact depth-first branch-and-bound dedicated to scheduling problems, using a *schedule or postpone strategy* as described by de Givry and Jeannin (2006) which avoids the enumeration of all possible starting time values. No initial upper bound was provided $(k = +\infty)$.

We generated 100 random instances for different numbers of candidate photographs ($N$ varying from 10 to 30)[2]. We compared BAC$^\varnothing$ (denoted by `BAC0` in the experimental results) with EDAC* (Heras et al., 2005) (denoted by `EDAC*`). Note that FDAC* and VAC (applied in preprocessing and during search, in addition to EDAC*) were also tested on these instances, but did not improve over EDAC* (FDAC* was slightly faster than EDAC* but developed more search nodes and VAC was significantly slower than EDAC*, without improving $w_\varnothing$ in preprocessing). OSAC is not practical on this benchmark (for $N = 20$, it has to solve a linear problem with $50,000$ variables and about 4 million constraints). All the algorithms are using the same search procedure. They are implemented in the `toulbar2` C++ solver[3]. Finding the minimum cost of the previously-described binary cost functions (which are convex if we consider the extra domain values for rejected photographs separately), is done in constant time for BAC$^\varnothing$. It is done in time $O(d^2)$ for EDAC* ($d = 201$).

We also report the results obtained by maintaining bounds consistency on the reified problem using meta-constraints as described by de Givry and Jeannin (2006), using the `claire/Eclair` C++ constraint programming solver (de Givry, Jeannin, Josset, Mattioli, Museux, & Savéant, 2002) developed by THALES (denoted by `B-consistency`).

The results are presented in Figure 4, using a log-scale. These results were obtained on a 3 GHz Intel Xeon with 4 GB of RAM. Figure 4 shows the mean CPU time in seconds and the mean number of backtracks performed by the search procedure to find the optimum and prove optimality as the problem size increases. In the legends, algorithms are sorted by increasing efficiency.

---

2. These instances are available at http://www.inra.fr/mia/ftp/T/bep/.

3. See http://carlit.toulouse.inra.fr/cgi-bin/awki.cgi/ToolBarIntro.

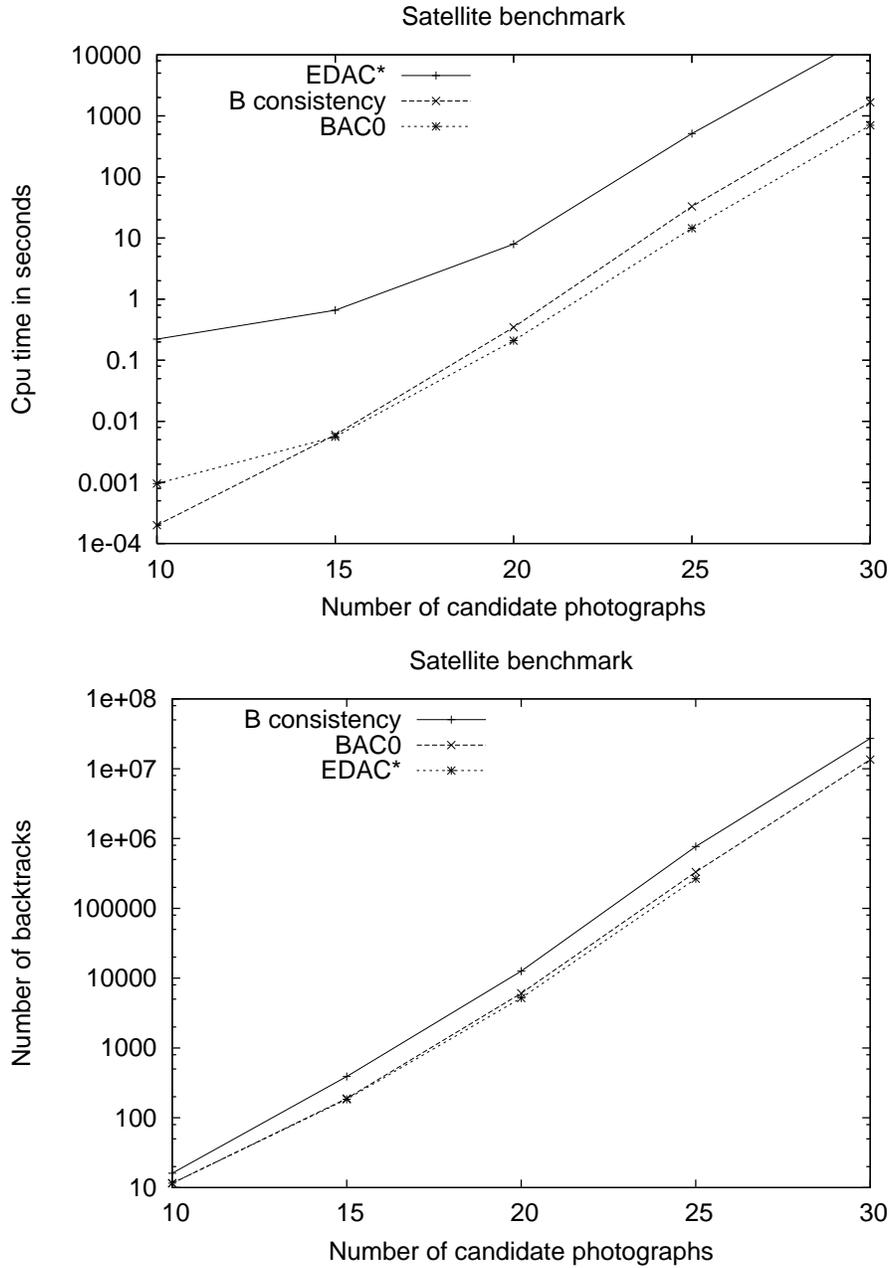

Figure 4: Comparing various local consistencies on a satellite benchmark. Cpu-time (top) and number of backtracks (bottom) are given.

The analysis of experimental results shows that BAC$^\varnothing$ was up to 35 times faster than EDAC* while doing only 25% more backtracks than EDAC* (for $N = 30$, no backtrack results are reported as EDAC* does not solve any instance within the time limit of 6 hours). It shows that bounds arc consistency can prune almost as many search nodes as a stronger local consistency does in much less time for temporal reasoning problems where the semantic of the cost functions can be exploited, as explained in Section 5. The second fastest approach was bounds consistency on the reified representation which was at least 2.3 worse than BAC$^\varnothing$ in terms of speed and number of backtracks when $N \geq 25$. This is a practical confirmation of the comparison of Section 6. The reified approach used with bounds consistency introduces Boolean decision variables for representing photograph selection and uses a criteria defined as a linear function of these variables. Contrarily to BAC$^\varnothing$, bounds consistency is by definition unable to reason simultaneously on the combination of several constraints to prune the starting times.

## 8.2 Non-coding RNA Gene Localization

A non-coding RNA (*ncRNA*) gene is a functional molecule composed of smaller molecules, called *nucleotides*, linked together by covalent bonds. There are four types of these nucleotides, commonly identified by a single letter: A, U, G and C. Thus, an RNA can be represented as a word built from the four letters. This sequence defines what is called the primary structure of the RNA molecule.

RNA molecules have the ability to fold back on themselves by developing interactions between nucleotides, forming pairs. The most frequently interacting pairs are: a G interacts with a C, or a U interacts with an A. A sequence of such interactions forms a structure called a *helix*. Helices are a fundamental structural element in ncRNA genes and are the basis for more complex structures. The set of interactions is often displayed by a graph where vertices represent nucleotides and edges represent either covalent bonds linking successive nucleotides (represented as plain lines in Figure 5) or interacting nucleotide pairs (represented as dotted lines). This representation is usually called the molecule's secondary structure. See the graph of a helix in Figure 5(a).

The set of ncRNAs that have a common biological function is called a *family*. The *signature* of a gene family is the set of conserved elements either in the sequence or the secondary structure. It can be expressed as a collection of properties that must be satisfied by a set of regions occurring on a sequence. Given the signature of a family, the problem we are interested in involves searching for new members of a gene family in existing genomes, where these members are in fact the set of regions appearing in the genome which satisfy the signature properties. Genomic sequences are themselves long texts composed of nucleotides. They can be thousand of nucleotides long for the simplest organisms up to several hundred million nucleotides for the more complex ones. The problem of searching for an occurrence of a gene signature in a genomic sequence is NP-complete for complex combinations of helix structures (Vialette, 2004).

In order to find ncRNAs, we can build a weighted constraint network that scans a genome, and detects the regions of the genome where the signature elements are present

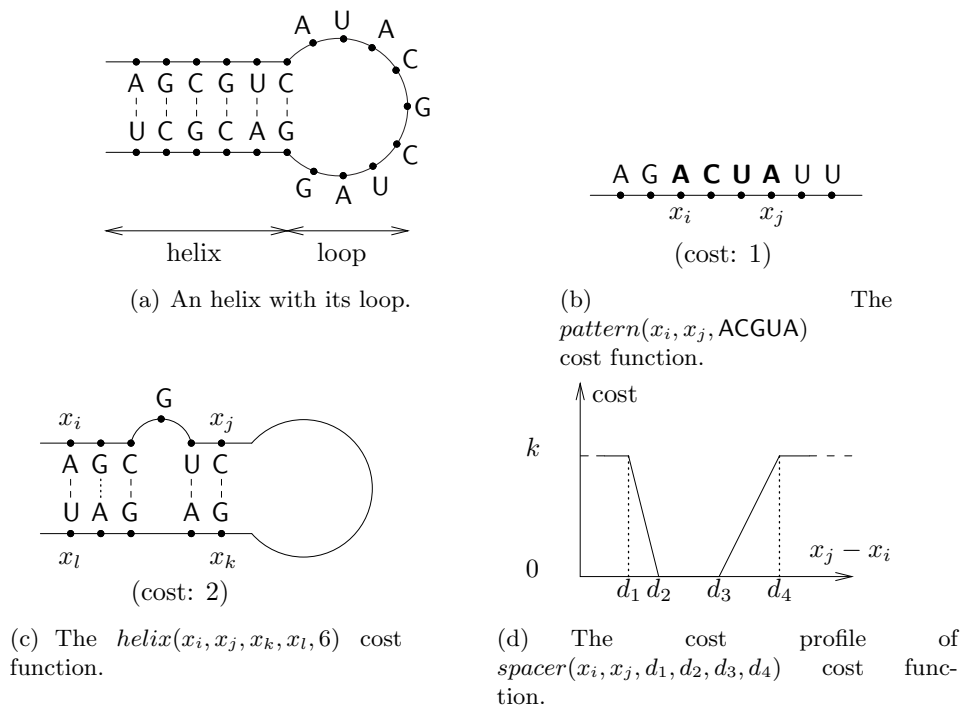

(a) An helix with its loop.

(b)      The $pattern(x_i, x_j, \mathsf{ACGUA})$ cost function.

(c) The $helix(x_i, x_j, x_k, x_l, 6)$ cost function.

(d)   The   cost   profile   of $spacer(x_i, x_j, d_1, d_2, d_3, d_4)$   cost   function.

Figure 5: Examples of signature elements with their cost functions.

and correctly positioned. The variables are the positions of the signature elements in the sequence. The size of the domains is the size of the genomic sequence. Cost functions enforce the presence of the signature elements between the positions taken by the variables involved. Examples of cost functions are given in Figure 5.

- The $pattern(x_i, x_j, p)$ function states that a fixed word $p$, given as parameter, should be found between the positions indicated by the variables $x_i$ and $x_j$. The cost given by the function is the edit distance between the word found at $x_i{:}x_j$ and the word $p$ (see the cost function $pattern$ with the word $\mathsf{ACGUA}$ in Figure 5(b)).

- The $helix(x_i, x_j, x_k, x_l, m)$ function states that the nucleotides between positions $x_i$ and $x_j$ should be able to bind with the nucleotides between $x_k$ and $x_l$. Parameter $m$ specifies a minimum helix length. The cost given is the number of mismatches or nucleotides left unmatched (see the $helix$ function with 5 interacting nucleotide pairs in Figure 5(c)).

- Finally, the function, $spacer(x_i, x_j, d_1, d_2, d_3, d_4)$ specifies a favorite range of distances between positions $x_i$ and $x_j$ using a trapezoidal cost function as shown in Figure 5(d).

See the work of Zytnicki et al. (2008) for a complete description of the cost functions.

Because of the sheer domain size, and given that the complex pattern matching oriented cost functions do not have any specific property that could speedup filtering, BAC alone has been used for filtering these cost functions (Zytnicki et al., 2008). The exception is the piecewise linear *spacer* cost function: its minimum can be computed in constant time for $\mathrm{BAC}^{\varnothing}$ enforcement. The resulting C++ solver is called `DARN!`[4].

| Size | 10k | 50k | 100k | 500k | 1M | 4.9M |
|------|-----|-----|------|------|-----|------|
| # of solutions | 32 | 33 | 33 | 33 | 41 | 274 |
| AC* | | | | | | |
| Time | 1hour 25min. | 44 hours | - | - | - | - |
| # of backtracks | 93 | 101 | - | - | - | - |
| BAC | | | | | | |
| Time (sec.) | 0.016 | 0.036 | 0.064 | 0.25 | 0.50 | 2.58 |
| # of backtracks | 93 | 101 | 102 | 137 | 223 | 1159 |

Table 1: Searching all the solutions of a tRNA motif in *Escherichia coli* genome.

A typical benchmark for the ncRNA localization problem is the transfer RNA (*tRNA*) localization. The tRNA signature (Gautheret, Major, & Cedergren, 1990) can be modelled by 22 variables, 3 nucleotide words, 4 helices, and 7 *spacers*. `DARN!` searched for all the solutions with a cost strictly lower than the maximum cost $k = 3$. Just to illustrate the absolute necessity of using bounds arc consistency in this problem, we compared bounds arc consistency enforcement with AC* (Larrosa, 2002) on sub-sequences of the genome of *Escherichia coli*, which is 4.9 million nucleotides long. Because of their identical space complexity and because they have not been defined nor implemented on non-binary cost functions (*helix* is a quaternarycost function), DAC, FDAC or EDAC have not been tested (see the work of Sànchez et al., 2008, however for an extension of FDAC to ternary cost functions).

The results are displayed in Table 1. For different beginning sub-sequences of the complete sequence, we report the size of the sub-sequence in which the signature is searched for (10k is a sequence of 10,000 nucleotides), as well as the number of solutions found. We also show the number of backtracks and the time spent on a 3 GHz Intel Xeon with 2 GB. A "-" means the instance could not be solved due to memory reasons, and despite memory optimizations. BAC solved the complete sequence in less than 3 seconds. BAC is approximately 300,000 (resp. 4,400,000) times faster than AC* for the 10k (resp. 50k) sub-sequence. More results on other genomes and ncRNA signatures can be found in the work of Zytnicki et al. (2008).

The reason of the superiority of BAC over AC* is twofold. First, AC* needs to store all the unary costs for every variable and projects costs from binary cost functions to unary cost functions. Thus, the space complexity of AC* is at least $\mathcal{O}(nd)$. For very large domains (in our experiments, greater than 100,000 values), the computer cannot allocate a sufficient memory and the program is aborted. For the same kind of projection, BAC only needs to store the costs of the bounds of the domains, leading to a space complexity of $\mathcal{O}(n)$.

Second, BAC does not care about the interior values and focuses on the bounds of the domains only. On the other hand, AC* projects all the binary costs to all the interior values, which takes a lot of time, but should remove more values and detect inconsistencies earlier. However, Table 1 shows that the number of backtracks performed by AC* and BAC are the same. This can be explained as follows. Due to the nature of the cost functions used in these problems, the supports of the bounds of the domains of the variables usually are the bounds of the other variables. Thus, removing the values which are inside the domains, as AC* does, do not help removing the bounds of the variables. As a consequence, the bounds founds by BAC are the same as those found by AC*. This explains why enforcing of AC* generally does not lead to new domain wipe out compared to BAC, and finding the support inside the bounds of the domains is useless.

Notice that the *spacer* cost functions dramatically reduce the size of the domains. When a single variable is assigned, all the other domain sizes are dramatically reduced, and the instance becomes quickly tractable. Moreover, the helix constraint has the extra knowledge of a maximum distance $d_{jk}$ between its variables $x_j$ and $x_k$ (see Fig. 5(c)) which bounds the time complexity of finding the minimum cost w.r.t. $d_{jk}$ and not the length of the sequence.

## 9. Conclusions and Future Work

We have presented here new local consistencies for weighted CSPs dedicated to large domains as well as algorithms to enforce these properties. The first local consistency, BAC, has a time complexity which can be easily reduced if the semantics of the cost function is appropriate. A possible enhancement of this property, $\varnothing$-IC, has also been presented. Our experiments showed that maintaining bounds arc consistency is much better than AC* for problems with large domains, such as ncRNA localization and scheduling for Earth observation satellites. This is due to the fact that AC* cannot handle problems with large domains, especially because of its high memory complexity, but also because $\text{BAC}^\varnothing$ behaves particularly well with specific classes of cost functions.

Similarly to bounds consistency, which is implemented on almost all state-of-the-art CSP solvers, this new local property has been implemented in the open source `toulbar2` WCSP solver.[5]

BAC, $\text{BAC}^\varnothing$ and $\varnothing$-inverse consistency allowed us to transfer bounds consistency CSP to weighted CSP, including improved propagation for specific classes of binary cost functions. Our implementation for RNA gene finding is also able to filter non-binary constraints. It would therefore be quite natural to try to define efficient algorithms for enforcing BAC, $\text{BAC}^\varnothing$ or $\varnothing$-inverse consistency on specific cost functions of arbitrary arity such as the soft global constraints derived from All-Diff, GCC or regular (Régin, 1994; Van Hoeve, Pesant, & Rousseau, 2006). This line of research has been recently explored by Lee and Leung (2009).

Finally, another interesting extension of this work would be to better exploit the connection between BAC and bounds consistency by exploiting the idea of Virtual Arc Consistency introduced by Cooper et al. (2008). The connection established by Virtual AC between crisp CNs and WCNs is much finer grained than in the reification approach considered by Petit et al. (2000) and could provide strong practical and theoretical results.

---

5. Available at http://carlit.toulouse.inra.fr/cgi-bin/awki.cgi/ToolBarIntro.

# References

Apt, K. (1999). The essence of constraint propagation. *Theoretical computer science*, *221*(1-2), 179–210.

Bessière, C., & Régin, J.-C. (2001). Refining the basic constraint propagation algorithm. In *Proc. of IJCAI'01*, pp. 309–315.

Chellappa, R., & Jain, A. (1993). *Markov Random Fields: Theory and Applications*. Academics Press.

Choi, C. W., Harvey, W., Lee, J. H. M., & Stuckey, P. J. (2006). Finite domain bounds consistency revisited. In *Proc. of Australian Conference on Artificial Intelligence*, pp. 49–58.

Cooper, M., & Schiex, T. (2004). Arc consistency for soft constraints. *Artificial Intelligence*, *154*, 199–227.

Cooper, M. C., de Givry, S., Sànchez, M., Schiex, T., & Zytnicki, M. (2008). Virtual arc consistency for weighted CSP.. In *Proc. of AAAI'2008*.

Cooper, M. C., de Givry, S., & Schiex, T. (2007). Optimal soft arc consistency. In *Proc. of IJCAI'07*, pp. 68–73.

de Givry, S., & Jeannin, L. (2006). A unified framework for partial and hybrid search methods in constraint programming. *Computer & Operations Research*, *33*(10), 2805–2833.

de Givry, S., Jeannin, L., Josset, F., Mattioli, J., Museux, N., & Savéant, P. (2002). *The THALES constraint programming framework for hard and soft real-time applications*. The PLANET Newsletter, Issue 5 ISSN 1610-0212, pages 5-7.

Freuder, E., & Wallace, R. (1992). Partial constraint satisfaction. *Artificial Intelligence*, *58*, 21–70.

Gautheret, D., Major, F., & Cedergren, R. (1990). Pattern searching/alignment with RNA primary and secondary structures: an effective descriptor for tRNA. *Comp. Appl. Biosc.*, *6*, 325–331.

Heras, F., Larrosa, J., de Givry, S., & Zytnicki, M. (2005). Existential arc consistency: Getting closer to full arc consistency in weighted CSPs. In *Proc. of IJCAI'05*, pp. 84–89.

Khatib, L., Morris, P., Morris, R., & Rossi, F. (2001). Temporal constraint reasoning with preferences. In *Proc. of IJCAI'01*, pp. 322–327.

Larrosa, J. (2002). Node and arc consistency in weighted CSP. In *Proc. of AAAI'02*, pp. 48–53.

Larrosa, J., & Schiex, T. (2004). Solving weighted CSP by maintaining arc-consistency. *Artificial Intelligence*, *159*(1-2), 1–26.

Lee, J., & Leung, K. (2009). Towards Efficient Consistency Enforcement for Global Constraints in Weighted Constraint Satisfaction. In *Proc. of IJCAI'09*.

Lhomme, O. (1993). Consistency techniques for numeric CSPs. In *Proc. of IJCAI'93*, pp. 232–238.

Meseguer, P., Rossi, F., & Schiex, T. (2006). Soft constraints. In Rossi, F., van Beek, P., & Walsh, T. (Eds.), *Handbook of constraint programming*, Foundations of Artificial Intelligence, chap. 9, pp. 281–328. Elsevier.

Petit, T., Régin, J.-C., & Bessière, C. (2000). Meta-constraints on violations for over constrained problems. In *Proc. of ICTAI'00*, pp. 358–365.

Petit, T., Régin, J. C., & Bessière, C. (2002). Range-based algorithm for Max-CSP. In *Proc. of CP'02*, pp. 280–294.

Régin, J.-C. (1994). A filtering algorithm for constraints of difference in CSPs. In *Proc. of AAAI'94*, pp. 362–367.

Sànchez, M., de Givry, S., & Schiex, T. (2008). Mendelian error detection in complex pedigrees using weighted constraint satisfaction techniques. *Constraints, 13*(1-2), 130–154.

Sandholm, T. (1999). An Algorithm for Optimal Winner Determination in Combinatorial Auctions. In *Proc. of IJCAI'99*, pp. 542–547.

Schiex, T. (2000). Arc consistency for soft constraints. In *Proc. of CP'00*, pp. 411–424.

Tarski, A. (1955). A lattice-theoretical fixpoint theorem and its applications. *Pacific Journal of Mathematics, 5*(2), 285–309.

Thébault, P., de Givry, S., Schiex, T., & Gaspin, C. (2006). Searching RNA motifs and their intermolecular contacts with constraint networks. *Bioinformatics, 22*(17), 2074–80.

Van Hentenryck, P., Deville, Y., & Teng, C.-M. (1992). A generic arc-consistency algorithm and its specializations. *Artificial Intelligence, 57*(2–3), 291–321.

Van Hoeve, W., Pesant, G., & Rousseau, L. (2006). On global warming: Flow-based soft global constraints. *Journal of Heuristics, 12*(4), 347–373.

Verfaillie, G., & Lemaître, M. (2001). Selecting and scheduling observations for agile satellites: some lessons from the constraint reasoning community point of view. In *Proc. of CP'01*, pp. 670–684.

Vialette, S. (2004). On the computational complexity of 2-interval pattern matching problems. *Theoretical Computer Science, 312*(2-3), 223–249.

Zytnicki, M., Gaspin, C., & Schiex, T. (2008). DARN! A soft constraint solver for RNA motif localization. *Constraints, 13*(1-2), 91–109.